\pdfoutput=1

\documentclass[11pt]{article}

\usepackage[final]{acl}
\newcommand{\mycomment}[1]{}
\usepackage{graphicx}
\usepackage{lscape}
\usepackage{tabularx}
\usepackage{longtable}
\usepackage{algorithm}
\usepackage{algpseudocode}
\usepackage{comment}

\usepackage{times}
\usepackage{latexsym}

\usepackage[T1]{fontenc}

\usepackage[utf8]{inputenc}

\usepackage{microtype}

\usepackage{inconsolata}

\title{How do you know that? Teaching Generative Language Models to Reference Answers to Biomedical Questions}

\author{Bojana Bašaragin \\
The Institute for AI of Serbia \\
Fruškogorska 1, Novi Sad \\
Serbia \\
 \texttt{bojana.basaragin@ivi.ac.rs} \\\And
Adela Ljajić \\
The Institute for AI of Serbia \\
Fruškogorska 1, Novi Sad \\
Serbia \\
 \texttt{adela.ljajic@ivi.ac.rs} \\\And
 Darija Medvecki \\
The Institute for AI of Serbia  \\
Fruškogorska 1, Novi Sad \\
Serbia \\
 \texttt{darija.medvecki@ivi.ac.rs} \\\AND
Lorenzo Cassano \\
Bayer A.G. \\
Müllerstraße 178, Berlin \\
Germany\\
 \texttt{lorenzo.cassano@bayer.com}\\\And
Miloš Košprdić \\
The Institute for AI of Serbia  \\
Fruškogorska 1, Novi Sad \\
Serbia \\
 \texttt{milos.kosprdic@ivi.ac.rs} \\\And
Nikola Milošević \\
Bayer A.G. \\
Müllerstraße 178, Berlin \\
Germany\\
 \texttt{ nikola.milosevic@bayer.com} \\}

\begin{document}
\maketitle
\begin{abstract}
Large language models (LLMs) have recently become the leading source of answers for users' questions online. Despite their ability to offer eloquent answers, their accuracy and reliability can pose a significant challenge. This is especially true for sensitive domains such as biomedicine, where there is a higher need for factually correct answers. This paper introduces a biomedical retrieval-augmented generation (RAG) system designed to enhance the reliability of generated responses. The system is based on a fine-tuned LLM for the referenced question-answering, where retrieved relevant abstracts from PubMed are passed to LLM's context as input through a prompt. Its output is an answer based on PubMed abstracts, where each statement is referenced accordingly, allowing the users to verify the answer. Our retrieval system achieves an absolute improvement of 23\% compared to the PubMed search engine. Based on the manual evaluation on a small sample, our fine-tuned LLM component achieves comparable results to GPT-4 Turbo in referencing relevant abstracts. We make the dataset used to fine-tune the models and the fine-tuned models based on Mistral-7B-instruct-v0.1 and v0.2 publicly available.

\end{abstract}

\section{Introduction}
The idea of automated referencing dates back to 1970 when \cite{garfield1970can} proposed an automatic system where a computer evaluates the appropriateness of references within an article. 
With the emergence of generative large language models (LLMs), numerous systems are being developed to answer specific questions, supported by relevant references \cite{huang2024citation,menick2022teaching,yang2023inference}.
Generative LLMs can produce answers that appear coherent, confident and articulate. However, the information conveyed may not be correct or verifiable. Furthermore, the limited internal knowledge of generative LLMs can hinder their ability to deliver factually accurate answers, particularly within specialized fields \cite{GRAVEL2023226,zheng2023does}. This issue is notably concerning in the biomedical domain, where accurate and factual answers are critical. The scientific community has recognized the dangers of factually incorrect or nonsensical information and has been reluctant to utilize these models to their potential. Providing an opportunity for scientists to obtain correct and verifiable answers to questions is an opportunity to increase scientific productivity and its impact. Moreover, privacy, sovereignty and security concerns in pharma and biomedicine often necessitate building systems where all components are controllable (e.g. deployed in-house), to avoid reliance on third-party APIs such as OpenAI\footnote{\url{https://openai.com}}, especially when secret data is concerned.

Incorporating domain-specific external knowledge beyond LLM data is essential for mitigating hallucinations in LLMs. The retrieval-augmented generation (RAG) approach, which integrates the generative capabilities of an LLM with a specialized retrieval system, enhances the model's accuracy and relevance by grounding its responses in verified information.

In this paper, we present a biomedical RAG system consisting of a hybrid search based on PubMed\footnote{\url{https://pubmed.ncbi.nlm.nih.gov}} and fine-tuned generative models for referenced question-answering (QA). We make both the models and the dataset used to fine-tune the models publicly available.

The remainder of this paper is organized as follows: Section \ref{related_work} provides a review of related work on reliability and verifiability of the LLM generated content and the approaches to generating texts with references. Section \ref{method} describes the design of the IR and generative components.  
We evaluate the components in Section \ref{results}, first individually and then jointly. We end the paper with conclusions and some future work remarks in Section \ref{future}.

\section{Related work} \label{related_work}
\begin{figure*}[h!]
     \centering
     \includegraphics[width=0.9\linewidth]{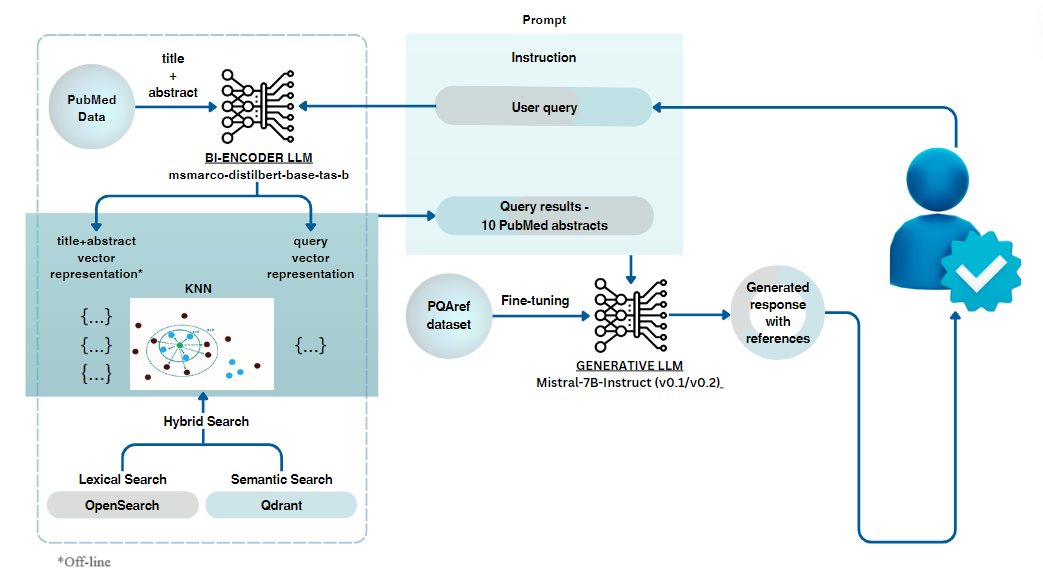}
     \caption{Architecture of our RAG system.}\label{Fig:system}
\end{figure*}

Generative LLMs, such as GPT and similar architectures, have enabled question-answering (QA) tasks across various domains, including medicine. The current state of these models is characterized by several challenges, particularly regarding the verifiability and reliability of the information they generate.
By evaluating ChatGPT responses and references in the medical domain, \cite{GRAVEL2023226} found that 69\% of generated references were fabricated, while professionals rated the answers at a median quality of 60\%. Similarly, when \cite{liu2023evaluating} conducted manual evaluations of four prominent generative search engines Bing Chat, NeevaAI, perplexity.ai, and YouChat, they found that while the responses of these engines were fluent and seemingly informative, only 51.5\% of sentences generated by these engines were fully supported by their citations, and merely 74.5\% of citations accurately supported the statements they were linked to. These results leave space for improvement.

In general, there are two approaches to generating text with references \cite{huang2024citation}. The first assumes training LLMs to produce references from parametric knowledge (information internalized from the training data). The second one assumes producing references from non-parametric knowledge (content retrieved from external sources).

The first approach, integrating citations directly from LLM's parametric knowledge, poses a significant technical challenge. Unlike search engines and IR systems that rely on indices for data retrieval, LLMs encode information into hidden representations during training, lacking a direct index. Therefore, referencing the sources of information becomes intricate. Despite these challenges, approaches have been suggested to train LLMs to include references using source identifiers \cite{taylor2022galactica}. However, these methods exhibit certain limitations, including citation inaccuracies and being restricted to academic citations.

The second approach, known as retrieval-augmented generation (RAG), combines generative LLMs with IR systems to form a hybrid system \cite{10.5555/3495724.3496517}. Here, the model is trained to recognize instances requiring citations, and the IR system retrieves suitable sources to provide context to the LLM. As a result, the LLM incorporates these sources as citations into its outputs, improving the credibility and accuracy of responses.
While pre-trained and fine-tuned LLMs rely solely on their parametric knowledge, RAG integrates a customized external knowledge base without additional training, thus reducing hallucinations. Moreover, annotators often perceive RAG-enhanced answers to be more factual and specific compared to those from fine-tuned models \cite{10.5555/3495724.3496517}.

\section{Method}\label{method}

The RAG system we propose in this paper is designed to perform referenced QA in the biomedical domain. It consists of two main components. The IR component, based on hybrid semantic and lexical search, retrieves relevant PubMed abstracts and provides a context for the generative LLM. The final system output is an answer to the user query, which contains a reference for each of the claims extracted from the relevant abstracts. The overview of the system architecture can be seen in Figure \ref{Fig:system}. 

\subsection{Information Retrieval Component}\label{IR}

Our IR component uses data from PubMed database\footnote{\url{https://pubmed.ncbi.nlm.nih.gov/download/}} containing citations and biomedical literature from several literature resources.
The IR system integrates both sparse vectors (lexical index) and dense vectors (semantic index), enabling lexical and semantic search, and a hybrid combination of the two.

For the lexical retrieval, based on a ranking function Best Matching 25 (BM25), we use the OpenSearch\footnote{\url{https://opensearch.org/}} to create an index for PubMed articles, by concatenation of title and abstract as an indexed field. Also, we add authors' names, publication dates, and journal names as metadata for filtering. 

For semantic retrieval, based on dense vectors, we use the Qdrant\footnote{\url{https://qdrant.tech/}} vector database. Qdrant allowed the usage of memory mapping of vectors to a hard drive, reducing the memory (RAM) requirements of the system.   
To optimize semantic search retrieval time, we used 8-bit quantized embeddings, with the option to use full embeddings for rescoring the results. 

We use the Hierarchical Navigable Small World (HNSW) indexing technique for Approximate Nearest Neighbors with dot product metrics to perform vector comparisons \cite{malkov2018efficient}. To create vector embeddings we use a bi-encoder sentence transformer model pre-trained on the MSMarco dataset \cite{Hofstaetter2021_tasb_dense_retrieval}, which, at the time of indexing, had the best performance on Passage Retrieval Task\footnote{\url{https://www.sbert.net/docs/pretrained-models/msmarco-v3.html}}.

In a corpus of 36,797,469 abstracts, 11,308,679 were found to be empty and thus omitted from the index. These empty abstracts predominantly originate from articles published in the pre-digital era, articles from journals that are not accessible for free, or journals that do not require abstracts. 
After eliminating these empty abstracts, we constructed two indices in the offline mode, designed for subsequent use in online semantic and lexical searches. 
The lexical index is created by indexing concatenated fields of titles and abstracts along with additional fields from PubMed articles for filtering purposes. The process of generating embeddings for the semantic index includes the creation of embeddings for titles and abstract concatenation using the model. This process is depicted in Figure \ref{Fig:system}, marked with an asterisk.
Before generating embeddings for semantic search, it was ascertained that the average number of tokens within the dataset's title and abstract concatenation was 650. Given that the maximum input size of the model employed for embedding creation is 512 tokens, abstracts exceeding this threshold were subdivided into segments each containing no more than 512 tokens, and were indexed separately. The split was made at the end of the sentence before the 512th token.

In our case, hybrid search is a combination of lexical and semantic IR components. To utilize the hybrid search, we normalized scores from these two IR methods to scales ranging from 0 to 1. The scores from each of the search methods are then multiplied by the importance weights for each of the methods. This allows both the identification of direct matches and greatly improves the ability to discover semantically related phrases and text segments, even in the absence of exact textual matches.
 
\subsection{Generative Component}\label{Gen component}

The generative component of our system is based on the Mistral-7B model. Despite having fewer parameters, Mistral-7B shows superior performance over larger models such as Llama 2 13B across all evaluated benchmarks and Llama 1 34B in reasoning benchmarks, maths, and code generation \cite{jiang2023mistral}. Compared to its 0.1 version, Mistral-7B v0.2 introduced an expanded context window (32K to the previous 8K) and several other adjustments (rope-theta = 1e6, no sliding-window attention) contributing to more accurate and consistent outputs, improved efficiency, and adaptability to many different tasks \cite{mistralv02}. 

For the sake of comparison, we opted for testing both currently available instruction-tuned versions of Mistral-7B (v0.1\footnote{\url{https://huggingface.co/mistralai/Mistral-7B-Instruct-v0.1}} and v0.2\footnote{\url{https://huggingface.co/mistralai/Mistral-7B-Instruct-v0.2}}). We test both models in the zero-shot mode but also fine-tune them using a custom dataset for referenced QA (see Section \ref{dataset-fine-tuning}).

The input for the generative component consists of a user query and 10 abstracts retrieved by the IR component as most relevant for the user query. While generating the answer, the models perform another relevance check and answer the question using only the abstracts they find relevant. The final output is a concise answer that contains an abstract ID as a reference after each claim originating from the 10 abstracts.

In the following subsections, we briefly describe the dataset we used to fine-tune these models, as well as the fine-tuning process.

\subsubsection{Dataset} \label{dataset-fine-tuning}

We created a custom dataset to fine-tune the LLMs for the task of referenced QA. The dataset consists of 9075 questions, where each question is followed by 10 relevant abstracts (along with titles and PMIDs) and referenced answers to the questions based on the provided abstracts.

The questions were randomly selected from the PubMedQA dataset \cite{jin2019pubmedqa}. The most relevant abstracts for each of these questions were retrieved from the PubMed repository using a combination of entity and free text search. To create the answers based on the retrieved abstracts, we used GPT-4 Turbo, specifically gpt-4-1106-preview\footnote{\url{https://platform.openai.com/docs/models/gpt-4-turbo-and-gpt-4}}, a GPT-4 Turbo preview model featuring improved instruction following. GPT-4 Turbo is currently the number one model on the Chatbot Arena leaderboard, a crowdsourced open platform for LLM evaluation \cite{chiang2024chatbot}. The prompt we used to instruct GPT-4 Turbo to use references (PMIDs) was as follows: 

\noindent
\qquad \\\fbox{
    \parbox{0.95\linewidth}{
Answer the question using relevant abstracts provided, up to 300 words. Reference the statements with the provided abstract\_id in brackets next to the statement.
       }
    }
\\

To ensure the completeness of answers, GPT-4 Turbo was further instructed to continue generating if there is more content to generate. The answers were then automatically checked for completeness and incomplete final sentences were removed, which finally led to the size of answers ranging from 69 to 1221 tokens. In a small number of cases (25 questions) there was no direct answer in the abstracts so the answer does not contain any references. The total input length in the dataset (question + abstracts + answer) ranges from 1686 to 6987 tokens.

We name this dataset PQAref and make it available through Hugging Face\footnote{\url{https://huggingface.co/datasets/BojanaBas/PQAref}}.

\begin{figure*}[ht]
    \centering
    \includegraphics[width=\textwidth]{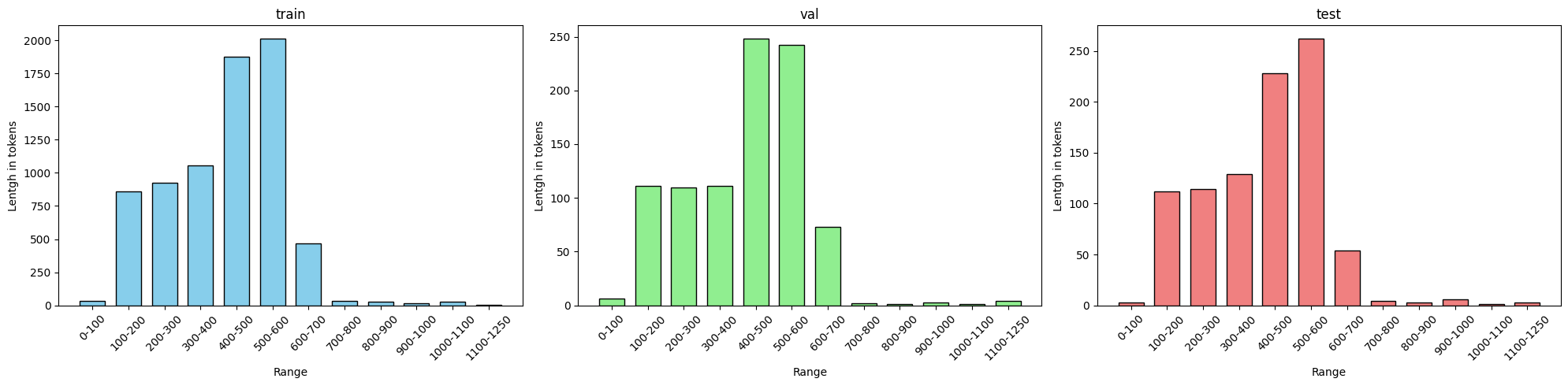}
    \caption{Distribution of answer length across train, val and test splits.}
    \label{fig:distribution}
\end{figure*}

\subsubsection{Fine-tuning the models}\label{Fine-tuning}

Both Mistral-7B instruction-tuned versions were fine-tuned for the task of referenced QA using the QLoRA methodology \cite{dettmers2023qlora}, allowing us to fine-tune the models on a single DGX NVIDIA A100-40GB GPU in $\sim$32 hours. The parameters we used for both models were standard loss, rank of 64, alpha of 16, and LoRA dropout of 0.1, resulting in 27,262,976 trainable parameters in both cases. Both models were fine-tuned over 2 epochs, using a batch size of 1. The PQAref dataset split was 80:10:10, with most inputs in the size range of 4000 to 6000 tokens in all three splits (see Figure \ref{fig:distribution}). 

We make the QLoRA adapters for both models available on Hugging Face as Mistral-7B-Instruct-v0.1-pqa-10\footnote{\url{https://huggingface.co/BojanaBas/Mistral-7B-Instruct-v0.1-pqa-10}} and Mistral-7B-Instruct-v0.2-pqa-10\footnote{\url{https://huggingface.co/BojanaBas/Mistral-7B-Instruct-v0.2-pqa-10}}.
\mycomment{
To obtain the answers from both fine-tuned models, we used the following prompt:

\noindent
\qquad \\\fbox{
    \parbox{0.95\linewidth}{
Respond to the Instruction using only the information provided in the relevant abstracts in ```Abstracts``` below.

\{instruction\}

Answer:
       }
    }
\\

The \textit{instruction} consists of the user query and 10 retrieved abstracts. We use default inference parameters for both models, except setting the repetition\_penalty to 1.1 and varying the values of
max\_new\_tokens. Despite adding the limit to the answers through the max\_new\_tokens parameter or through trying to add a limit to the prompt (e.g. "Answer in at most 300 words."), both models continuously generated an arbitrary number of tokens. The same behavior was noticed in GPT-4 Turbo during the creation of the PQAref dataset. The token limitation, primarily imposed due to the prolonged inference time for higher values (e.g. 1000) would only lead to interrupted answers. Finally, the limit was set to 1225, to match the longest complete answer length in the training dataset. An example of all three models' answers to an instruction from the test set can be seen in Appendix \ref{tab:appendixA}.
}

\section{Results} \label{results}

\subsection{Evaluation of IR Component}\label{IR evaluation}

To evaluate our IR system, we utilized the BioASQ dataset \cite{bioasq_dataset}. 
The BioASQ dataset is designed for tasks that help drive advancements in biomedical information retrieval and QA. It includes 5049 questions along with corresponding gold-standard answers, relevant document snippets, and the PubMed IDs (PMIDs) of articles that are relevant to each question.

We compared the PMIDs retrieved by our system against the gold-standard PMIDs provided in the BioASQ dataset. This comparison was quantified using the precision metric, measuring the proportion of relevant identifiers retrieved by our system out of the total PMIDs retrieved. We evaluate precision using 10 retrieved documents (P@10) and mean average precision for 10 retrieved documents (MAP@10). The evaluation of the retrieval component is done using: (1) only lexical, (2) only semantic, and (3) a combination of the two. Additionally, we experimented with different weights for the lexical and semantic combinations. 

For the lexical search, we experimented with stopword removal from the query and obtained better results compared to lexical search without stopword removal as shown in Table \ref{tab_IR_performance}.

For semantic search, we experimented with three approaches: semantic search with full embeddings, semantic search with compressed embeddings (using 8-bit quantization), and semantic search using compressed embeddings with rescoring (using full embeddings for rescoring).

Semantic search with full embeddings had an average response time of 30 seconds, making it inefficient and unusable for real-world applications.

For semantic search with rescoring, we used compressed embeddings to retrieve 100 results, then rescored the top 10 using full-size embeddings. This method improved precision by 0.3\% and was only 52 milliseconds slower than the approach without rescoring (see rows 1 and 2 in Table \ref{tab_IR_performance}). Given the minimal additional time required,  
 we tested the various weight combinations of hybrid search incorporating semantic search with rescoring. Parallel execution of semantic and lexical search further contributes to the time efficacy of the system (as shown in Table \ref{tab_IR_performance}), reducing the average execution time from 489ms to 442ms. 
 
\begin{table}[h]
    \centering
    \caption{Our IR and PubMed search engine performance evaluation on the BioASQ dataset.
    }
    \label{tab_IR_performance}
    \resizebox{\columnwidth}{!}{%
    \begin{tabular}{lcccc}
\hline
         & P@10 & MAP@10 & time [ms] \\
\hline
        1. Semantic without rescore & 14.0\% & 25.7\%  & 245  \\
        2. Semantic with rescore & 14.4\% & 26.0\% & 297\\
        3. Hybrid with rescore (lex. 0.1 sem. 0.9) & 24.7\% & 32.5\% & 442\\
        4. Hybrid with rescore (lex. 0.2 sem. 0.8) & 24.7\% & 32.5\% & 442\\
        5. Hybrid with rescore (lex. 0.3 sem. 0.7) & 24.7\% & 32.5\% & 442\\
        6. Hybrid with rescore (lex. 0.4 sem. 0.6) & 24.7\% & 32.6\% &  442\\
        7. Hybrid with rescore (lex. 0.5 sem. 0.5) & 25.2\% & 41.0\% & 442\\
        8. Hybrid with rescore (lex. 0.6 sem. 0.4) & 30.7\% & 42.0\% & 442\\
        \textbf{9. Hybrid with rescore (lex. 0.7 sem. 0.3)}  &\textbf{ 30.8\%} & \textbf{42.5\%} & \textbf{442}\\
        10. Hybrid with rescore (lex. 0.8 sem. 0.2)  & 30.8\% & 42.5\% & 442 \\
        11. Hybrid with rescore (lex. 0.9 sem. 0.1)  &	30.8\% &42.6\% & 442\\
        12. Lexical with stopwords removal & 28.7\% & 41.1\%  & 189 \\
        13. Lexical without stopwords removal & 28.3\% & 40.1\%  & 189 \\

\hline
        14. PubMed without MeSH Terms	& 9.2\% & 	15.3\% & 698 \\
        15. PubMed with MeSH Terms &	12.0\% &	19.1\% & 742	\\
\hline

    \end{tabular}%
}
\end{table}

From the experiments detailed in Table \ref{tab_IR_performance}, it is evident that the performance of semantic search alone is suboptimal, with notable enhancements observed upon integration with lexical search. The initial improvement is noted with the hybrid search employing a 0.1 lexical search weight, followed by a second significant enhancement achieved with a 0.6 lexical search weight (yielding absolute improvements of 10.3\% and 16.3\% respectively). Increasing the lexical search weight beyond 0.6 does not yield noticeably different outcomes. Assigning a weight of 1 to lexical search in hybrid search excludes semantic search, effectively reducing the system to pure lexical search, which produces worse results.

As the subsequent generative component does not account for the order of retrieved documents, we employ the P@10 metric to determine the most effective combination of parameters for hybrid search. After evaluating various configurations, we identified the optimal parameters for hybrid search: a lexical search weight of 0.7 and a semantic component weight of 0.3. By allocating a higher weight to the semantic search component (0.3 in row 9 instead of 0.1 in row 11), we enhance the model's ability to capture and utilize the deeper, contextual relationships inherent in biomedical texts. Consequently, as shown in row 9, we choose these parameter values to conduct a hybrid search in our system.

Additionally, we evaluated the performance of PubMed search engine on the BioASQ dataset and got the P@10 of 9.2\% and MAP@10 of 15,3\% when searching without MeSH terms and P@10 of 12\% and MAP@10 of 19.1\% when searching with MeSH terms (rows 14 and 15 in Table \ref{tab_IR_performance}). 

\subsection{Evaluation of the Generative Component}\label{eval_generativ}

For the purpose of standalone evaluation of the generative component, we use the PQAref test set. We conducted automated and manual evaluations for the task of referenced QA, which involved analyzing the total number of all references per answer and relevant references per answer, checking the correctness of IDs, and comparing the number of relevant references to irrelevant ones in the models' answers.

To obtain the referenced answers in the zero-shot mode, we opted for the following prompt:

\noindent
\qquad \\\fbox{
    \parbox{0.95\linewidth}{
Respond to the Instruction using only the information provided in the relevant abstracts under Abstracts. Reference the statements with the provided abstract\_id in brackets next to the statement (for example PUBMED:1235):\\\{instruction\}
       }
    }
\\

To obtain the referenced answers from the fine-tuned models, we use the following prompt:

\noindent
\qquad \\\fbox{
    \parbox{0.95\linewidth}{
Respond to the Instruction using only the information provided in the relevant abstracts in ```Abstracts``` below.

\{instruction\}
       }
    }
\\

Both prompts were chosen after extensive testing of several different prompting strategies and prompt versions.

We use default inference parameters for all four models, except setting the repetition\_penalty to 1.1 for the fine-tuned models and varying the values of
max\_new\_tokens (max\_tokens for the zero-shot mode) for all four models. Despite trying to add the limit to the answers through the max\_new\_tokens parameter or through trying to add a limit to the prompt (e.g. "Answer in at most 300 words."), all the models continuously generated an arbitrary number of tokens. The same behavior was noticed in GPT-4 Turbo during the creation of the PQAref dataset. Token limitation, primarily imposed due to the prolonged inference time for higher values, often led to interrupted answers. Finally, the limit was set to 1225, to slightly exceed the longest complete answer length in the training dataset (see Section \ref{dataset-fine-tuning}). 

We refer to the zero-shot results of these two models as 0-M1 for v0.1 and 0-M2 for v0.2 and to the results of the fine-tuned models as M1 for v0.1 and M2 for v0.2. In both prompts, the \textit{instruction} for the fine-tuned models consists of the user query and 10 retrieved abstracts. An example of a question and GPT-4 Turbo's answer from the test set, along with other four models' answers to the same question can be seen in Appendix \ref{tab:appendixA}. 

\textbf{Automated evaluation.} The number of referenced abstracts in generated answers within PQAref test set (containing 908 examples) can be seen in Table \ref{references}. What can be observed is that 1 reference per answer is most common in GPT-4 Turbo answers from PQAref (241 answers). M1 and M2 have the highest number of answers with 3 references (185 cases for M1 and 178 for M2). In the case of zero-shot results, both 0-M1 and 0-M2 most commonly did not reference any abstracts in their responses: 527 occurrences (58\% of all the answers) for 0-M1 and 165 for 0-M2 (18.2\% of all the answers). M1 and M2 did not reference any abstracts in 8 (0.9\%) and 5 (0.5\%) answers, respectively. By manual inspection of these answers, the models stated that none of the abstracts were relevant, demonstrating their proficiency in task execution. On the other hand, in most of the answers without references 0-M1 and 0-M2 answered the question but without providing any references to their statements. Additionally, some 0-M2's answers (35 of them) repeated the first part of the instruction, suggesting the need for further postprocessing of its answers.

\begin{table}[h]
\caption{Number of referenced abstracts per model on the PQAref test set. N: number of referenced abstracts per answer. TOTAL: is the sum of referenced abstracts per model. AVG: the average number of references per answer.}\label{references}
\centering
\resizebox{\columnwidth}{!}{%
\begin{tabular}{lccccc}
 & \multicolumn{5}{c}{Number of answers containing N references}\\
\hline
N                               & GPT-4 Turbo  & 0-M1 & 0-M2 & M1           & M2           \\ \hline
0                               & 2            & \textbf{527}             & \textbf{165}             & 8            & 5            \\
1                               & \textbf{241} & 27                       & 11                       & 86           & 105          \\
2                               & 76           & 66                       & 47                       & 138          & 112          \\
3                               & 128          & 28                       & 92                       & \textbf{185} & \textbf{178} \\
4                               & 126          & 17                       & 114                      & 172          & 169          \\
5                               & 119          & 25                       & 110                      & 117          & 124          \\
6                               & 87           & 28                       & 94                       & 72           & 75           \\
7                               & 45           & 26                       & 61                       & 66           & 34           \\
8                               & 29           & 47                       & 64                       & 27           & 34           \\
9                               & 31           & 47                       & 83                       & 22           & 23           \\
10                              & 24           & 70                       & 67                       & 15           & 49           \\ \hline
TOTAL     & 3,464        & 2,285                        & 4,307                        & 3,648        & 3,816        \\ \hline

AVG & 3.81         & 2.51                        & 4.74                        & 4.01         & 4.20         \\
\end{tabular}
}
\end{table}

\begin{table*}[t]
    \centering
    \caption{The number of missed and referenced most relevant abstracts of 823 abstracts across the models.}
    \label{tab:}
    \resizebox{\textwidth}{!}{%
    \begin{tabular}{lccccc}
\hline
         & GPT-4 Turbo & 0-M1 & 0-M2 & M1 & M2\\
\hline
        Relevant missed & 1 (0.1\%) & 497 (60.4\%) & 185 (22.5\%) & 29 (3.5\%) & 10 (1.2\%)\\
        Relevant referenced & 822 (99.9\%) & 326 (39.6\%) & 638 (77.5\%) & 794 (96.5\%) & 813 (98.8\%) \\
\hline
    \end{tabular}%
}
\end{table*}

In the entire test set, comprising 908 examples with a total of 9080 abstracts (10 abstracts per example), 0-M2 has the highest average number of references per answer of 4.74, followed by M2 with 4.2 and M1 with 4.01, while 0-M1 produced 2.51 references per answer.

To measure the relevance of the referenced abstracts, we evaluated whether the models referenced at least the most relevant abstract for each question. Our dataset contains questions from Pub-MedQA, which in a number of cases originate from actual PubMed abstract titles. This means that during retrieval, the article whose title matches the question is very likely to be retrieved as relevant. In our test split, this indeed is the case in 823 out of 908 inputs. We decided to take such abstracts as the most relevant ones for those 823 inputs, which allowed us to automatically measure the number of times the models referenced that particular abstract. Table \ref{tab:} presents the number of missed and referenced most relevant abstracts using this tactic. When looking at the GPT-4 Turbo answers from the test set, the most relevant article was missed in only one case, suggesting it served as a good referencing role model. M2 missed the relevant abstract in 10 examples, while M1 missed it in 29 examples. Overall, both fine-tuned models do reference the most relevant abstract in most cases (96.5\% and 98.8\% respectively). On the other hand, 0-M1 missed the most relevant abstracts in 60.4\% answers and 0-M2 in 22.5\% answers, which shows a significantly weaker ability of the models to identify and extract the most relevant abstracts compared to their fine-tuned versions.

We also evaluated whether all the IDs in the models' answers matched the PMIDs of context-provided abstracts to verify none of them were hallucinated. GPT-4 Turbo's answers in the PQAref dataset contained no hallucinated IDs. However, both M1 and M2 produced hallucinated IDs, with a notable discrepancy. M1 produced 79 hallucinated IDs, while M2 produced only 3. The hallucinated IDs differ from the actual IDs by one or two digits. Upon manual inspection of the answer content and referenced IDs, we found that M1 tended to blend information from various abstracts, whereas M2 utilized information solely from the corresponding abstract. This suggests that M2 exclusively hallucinated some of the digits from the existing abstract ID, but not the content. This behavior remains consistent across different temperature values of the model. Looking at the zero-shot performance, 0-M1 hallucinated 11 IDs. However, it also did not reference any abstracts in 58\% of cases, which then presents an even higher number compared to the number of answers containing references. 0-M2 hallucinated in case of 26 IDs. The results point to a clear advantage of M2's answers in this respect.

\textbf{Manual evaluation.} To perform manual evaluation, we extracted 10 random examples from the PQAref test set. We then manually assessed the relevance of each of the abstracts in the examples. We generally distinguished between two types of abstracts: relevant and irrelevant. The abstracts we considered \textit{relevant} were the ones that covered all the specific aspects of the question and thus provided direct answers. Among them,  we defined the abstracts whose title matched the question as \textit{the most relevant} (as mentioned for 823 examples during automatic evaluation).
On the other hand, we identified two types of \textit{irrelevant} abstracts. The first type includes abstracts that miss the main topic of the question (e.g. discuss heart failure instead of knee problems), which we considered \textit{completely irrelevant} abstracts. The other type that discusses a more general topic and thus does not cover all the aspects of the question we considered \textit{partially irrelevant}. This group could also be observed as the one that contains additional information but does not provide the direct answer to the question. 

It is crucial to recognize that there can be two types of mistakes when irrelevant abstracts are concerned. If the model references a completely irrelevant abstract that is a clear mistake, however, if it references a partially irrelevant abstract, whether it is wrong may depend on the other references in the answer. If the answer also contains the reference that gives a direct answer to the question (relevant abstract), this could be considered additional information. If this is not the case, the model may have missed the main point. 

Finally, we examined how the models referenced the most relevant and irrelevant abstracts. For these 10 qualitatively observed examples, the fine-tuned models referenced the most relevant abstracts every time, meaning that they grasped the main point. On the other hand, 0-M1 and 0-M2 failed to reference these abstracts 4 and 2 times. Moreover, these answers of 0-M1 and 0-M2 contained no references whatsoever. None of the models referenced completely irrelevant abstracts. The general tendency of all four models was to provide additional information by referencing partially irrelevant abstracts. In several situations, the models seemed to filter the abstracts based on their understanding of a term used in the question, thus excluding the abstracts that use a different phrasing or an extended meaning of the term (e.g. donation taken to refer only to organ, tissue or bone marrow donation and not to cell and blood donation). 

We also conducted a quantitative analysis to examine how well they identified all the relevant abstracts. To overcome variations in the number of relevant abstracts per document and document-specific characteristics, we considered all 100 abstracts, 10 for each of 10 questions, collectively.
 
Of these 100 abstracts, the evaluators identified 42 relevant and 58 irrelevant abstracts. We prioritized and calculated recall for relevant abstracts for each model, as our primary concern is their ability to correctly identify and reference relevant abstracts. M1 exhibited the highest recall of 0.76, followed by M2 with 0.67, 0-M2 with 0.62 and 0-M1 with 0.29. For reference, the recall measured on the GPT-4 Turbo answers from the test set totalled 0.62. These results are summed up in the first row of Table \ref{recall}. The findings suggest that, based on the analysis of these 10 manually reviewed documents, M1 outperforms the other models in terms of referencing abstracts deemed relevant by evaluators, showing the highest benefit from the fine-tuning process.

\begin{table}[h]
    \centering
    \caption{Recall values for relevant abstracts on 10 examples from the PQAref test set and same 10 questions with abstracts retrieved with our IR system.}
    \label{recall}
    \resizebox{\columnwidth}{!}{%
    \begin{tabular}{lccccc}
\hline
    & GPT-4 Turbo & 0-M1 & 0-M2 & M1 & M2\\
\hline
        PQAref & 0.62 & 0.29 & 0.62 & \textbf{0.76} & \underline{0.67}\\
        IR  & 0.46 & 0.37 & \underline{0.59} & \textbf{0.64} & 0.58 \\
\hline
    \end{tabular}%
}
\end{table}

\subsection{System evaluation}\label{Joint eval}

In this section, we provide the preliminary joint evaluation of our system: the IR component (based on hybrid lexical and semantic search) and the generative component using the outputs of our IR, 

We manually evaluated the IR output on the same 10 PQAref questions we chose for the evaluation of the generative component in Section \ref{eval_generativ}. To retrieve the relevant abstract from indexed PubMed articles, we utilized the best-performing hybrid search parameter combination from Section \ref{IR evaluation} and retrieved 10 abstracts for each question. After manually determining the abstract relevance, we obtained 50\% P@10. This metric underscores the effectiveness of our IR component in locating documents for query responses. The fact that IR evaluation on BioASQ reached the best performance of P@10 30.8\% with the same combination of weights for hybrid search as manual evaluation on PQAref, further corroborates the results obtained in manual evaluation conducted on the PQAref dataset. 

We then used the same prompt for GPT-4 Turbo as in Section \ref{dataset-fine-tuning}, and the ones used in Section \ref{eval_generativ} for 0-M1, 0-M2, M1 and M2, to generate referenced answers based on the retrieved documents. We further computed the recall values for the relevant abstracts in the 10 generated answers and displayed them in the second row of Table \ref{recall}. It is noticeable that, once again, the model that performed best is M1, with the recall of 0.64. This model cites a greater number of abstracts that contain the relevant answers compared to other models. Based solely on the recall, 0-M2 showed better results compared to M2, albeit by only 0.01. However, in one of 10 examples it did not provide any references to its elaborate answer. M2, as the third best model with recall of 0.58 properly referenced all the answers. From Table \ref{references}, we can also observe that the model with most references is 0-M2, but it also does not provide any references in 18.2\% of the answers. Taking this important aspect into consideration, M2's answers prove more reliable compared to 0-M2. M2 shows a slightly lower recall compared to M1 because it has fewer references to abstracts that provide direct answers to the questions. Nonetheless, since the IR component consistently finds documents related to the topic, we give preference to M2's answers since they include more additional citations, offering more elaborate answers on the same topics. Here, GPT-4 Turbo had the recall of 0.46, while 0-M1 had the lowest recall of all the models (0.37), owing to a large number of answers with no references (5 out of 10).

\section{Conclusions and future work} \label{future}

In this paper, we provide an overview of biomedical generative search with answers grounded in PubMed and referenced claims. Our aim was to develop a system capable of generating accurate and verifiable answers to biomedical questions while maintaining user sovereignty and leveraging open-source models.

Starting with our IR component, we discovered that employing a combination of lexical and semantic searches yields the highest precision score. Our system demonstrates an absolute improvement of 23.4\% MAP@10 measure compared to the PubMed search engine. Through separate evaluations, we found that lexical search alone outperforms semantic search. However, integrating both approaches is advantageous for identifying instances lacking exact term matches, where semantic search contributes significantly. To enhance semantic search performance in IR, one future direction is to fine-tune these models on domain-specific data. This approach aims to improve the quality of embeddings in the biomedical domain, enabling them to encode domain-specific knowledge better, enhance contextual understanding, and ultimately improve IR performance.

Overall, the Mistral 7B Instruct models performed comparatively to GPT-4 Turbo in terms of the task of referenced QA. Based on the evaluation of the whole PQAref test set, M1 and M2 showed superior performance over 0-M1 and 0-M2 in referencing the most relevant abstracts, with M2 showing an improved performance of 2.3\% over M1, 21.3\% over 0-M2 and 59.2\% over 0-M1. As a general trend, M2 includes more information in its answers.

All four models showed hallucinations when generating IDs of references. Once again, M2 performed best in this respect with only 3 mismatches in ID digits, followed by 0-M1 (11) and 0-M2 (26), with the worst performance of 79 hallucinated answers coming from M1. While M2 was still using correct information from the corresponding abstract, this point needs further attention. Exchanging the IDs with numerals (1-10) for each abstract during fine-tuning could potentially solve this issue. This is something we plan to try in the next iteration of the dataset and training.

In terms of recall values for relevant abstracts, based on the manual evaluation of 10 examples from PQAref test set both fine-tuned models performed better, exhibiting a 47\% and 5\% improvement over their versions in zero-shot mode. The situation is slightly different for the same 10 questions with abstracts retrieved using our IR. While recall values of M1 were still superior, especially compared to 0-M1 with 27\% improvement, 0-M2 performed slightly better than M2 (1\% difference). However, considering the number of hallucinations and answers with no references, we give clear preference to the fine-tuned models.

It is worth noting that on a small test set, we have achieved a comparable and sometimes even better performance than GPT-4 Turbo with much smaller, open-source and fine-tuned models. Having an automated evaluation method of referenced QA quality would allow us to make a more comprehensive comparison of our models to models such as GPT-4 Turbo. To supplement manual evaluation and speed up the evaluation process, \cite{gao2023enabling} created a benchmark for automatic evaluation of citation quality and proposed an entailment model as a method for automatic quality check of references. We intend to further check the quality of answers on a larger sample using the same method. At this point, M2 gives more reliable answers, but the decision on which model to use in our final pipeline will be made after this evaluation.

\section*{Acknowledgments}

This work is funded within the framework of the NGI Search project under grant agreement No 101069364. The resources for the fine-tuning process were provided to us by the National Platform for Artificial Intelligence of the Republic of Serbia. More information and live demo can be found at \url{https://verifai-project.com}.

\clearpage

\bibliography{custom}

\clearpage

\appendix

\label{sec:appendix}
\onecolumn
\section*{Appendix A: Example answers of generative LLMs}
\setcounter{table}{0}
\renewcommand{\thetable}{A\arabic{table}}

\begin{table*}[ht]

\caption{Models' answers to the question "Does leisure time physical activity in early pregnancy protect against pre-eclampsia?"}
\label{tab:appendixA}
\begin{tabular}{p{\textwidth}}
\hline
\small{GPT-4 Turbo} \\ 
\hline
\small{The relationship between leisure time physical activity (LTPA) in early pregnancy and the risk of pre-eclampsia is not clearly established, with studies showing mixed results. Some studies suggest that there may be a protective effect of LTPA against pre-eclampsia, while others do not find a significant association or suggest an increased risk with high levels of activity.

A prospective cohort study in Denmark found that moderate levels of physical activity (1-270 minutes/week) in the first trimester had no statistically significant association with the risk of pre-eclampsia, but high levels of activity (exceeding 270 minutes/week) were associated with an increased risk of severe pre-eclampsia (PUBMED:19055653). Similarly, a population-based prospective cohort study in Amsterdam reported that the amount of time or intensity of physical activity in leisure time was not associated with a difference in risk of pre-eclampsia or gestational hypertension (PUBMED:19916877).

In contrast, a study from the New Haven, Connecticut area suggested that women who engaged in any regular LTPA during early pregnancy were at decreased risk of pre-eclampsia (PUBMED:15466498). Another study found a tendency towards a lower risk of pre-eclampsia in women with the highest degree of physical activity during leisure time, especially in overweight women, although no significant associations were found (PUBMED:20121498).

A systematic review examining the association between LTPA before and/or during pregnancy and the risk of pre-eclampsia found mixed results, with some studies suggesting that vigorous-intensity LTPA may reduce the risk of pre-eclampsia, while one study reported an elevated risk with high amounts of LTPA (PUBMED:23836014). Another case-control study found that women who performed regular LTPA during the first 20 weeks of pregnancy had a reduced risk of pre-eclampsia and gestational hypertension (PUBMED:2592903).

In summary, the evidence is inconclusive regarding the protective effect of LTPA in early pregnancy against pre-eclampsia. While some studies suggest a potential benefit, especially with moderate activity levels, others indicate no effect or an increased risk with high levels of activity. It is important to consider the intensity and duration of LTPA when evaluating its impact on pre-eclampsia risk.} \\
\hline
 \small{0-M1}\\
 \hline
  \small{The instruction is to provide a response to the instruction using only the information provided in the abstracts. The instruction is: "Does leisure time physical activity in early pregnancy protect against pre-eclampsia?"

Based on the provided abstracts, the answer is: No, leisure time physical activity in early pregnancy does not protect against pre-eclampsia.}
  \small{0-M2} \\
  \hline
   \small{The available evidence from the provided abstracts suggests that the relationship between leisure time physical activity (LTPA) in early pregnancy and the risk of pre-eclampsia is not clear-cut. Some studies indicate a protective effect of LTPA against pre-eclampsia, while others do not find a significant association or suggest an increased risk with high levels of activity.

One prospective cohort study in Denmark found that moderate levels of physical activity (1-270 minutes/week) in the first trimester had no statistically significant association with the risk of pre-eclampsia, but high levels of activity (exceeding 270 minutes/week) were associated with an increased risk of severe pre-eclampsia (PUBMED:19055653). Another population-based prospective cohort study in Amsterdam reported that the amount of time or intensity of physical activity in leisure time was not associated with a difference in risk of pre-eclampsia or gestational hypertension (PUBMED:19916877).

However, a study from the New Haven, Connecticut area suggested that women who engaged in any regular LTPA during early pregnancy were at decreased risk of pre-eclampsia (PUBMED:15466498). Another study found a tendency towards a lower risk of pre-eclampsia in women with the highest degree of physical activity during leisure time, especially in overweight women, although no significant associations were found (PUBMED:20121498).

A systematic review examining the association between LTPA before and/or during pregnancy and the risk of pre-eclampsia found mixed results, with some studies suggesting that vigorous-intensity LTPA may reduce the risk of pre-eclampsia, while one study reported an elevated risk with high amounts of LTPA (PUBMED:23836014). Another case-control study found that women who performed regular LTPA during the first 20 weeks of pregnancy had a reduced risk of pre-eclampsia and gestational hypertension (PUBMED:2592903).

In summary, the evidence is inconclusive regarding the protective effect of LTPA in early pregnancy against pre-eclampsia. While some studies suggest a potential benefit, especially with moderate activity levels, others indicate no effect or an increased risk with high levels of activity. It is important to consider the intensity and duration of LTPA when evaluating its impact on pre-eclampsia risk. Further research is needed to clarify the relationship between LTPA and pre-eclampsia risk.}
\end{tabular}

\end{table*}

\clearpage
\begin{table*}[htbp]
\begin{tabular}{p{\textwidth}}
\hline
  \small{M1} \\
\hline
   \small{Based on the provided abstracts, leisure time physical activity in early pregnancy does not appear to protect against pre-eclampsia. A prospective cohort study in Denmark found that the two highest levels of physical activity were associated with increased risk of severe pre-eclampsia compared with the nonexercising group, suggesting that leisure time physical activity exceeding 270 minutes/week in first trimester may increase the risk of severe pre-eclampsia (PUBMED:19055653). 
  
  Another study in Amsterdam, The Netherlands, did not find any association between physical activity in leisure time early in pregnancy and the incidence of preeclampsia or gestational hypertension (PUBMED:19916877). Similarly, a study in New Haven, Connecticut, found that women who engaged in any regular leisure time physical activity, regardless of caloric expenditure, were at decreased risk of preeclampsia, but there was no indication of a protective effect of workplace activity, leisure-time physical activity, or unemployment (PUBMED:15466498). These findings suggest that while certain forms of physical activity may have a protective effect on preeclampsia, the overall evidence does not support a clear protective effect of leisure time physical activity in early pregnancy against this condition. Therefore, based on the available abstracts, leisure time physical activity in early pregnancy does not protect against pre-eclampsia.} \\
\hline
 \small{M2} \\
\hline
\small{Based on the provided abstracts, there is conflicting evidence regarding whether leisure time physical activity (LTPA) in early pregnancy protects against pre-eclampsia. Some studies suggest that LTPA may offer protection against pre-eclampsia, while others do not find a significant association or even indicate that excessive LTPA could increase the risk of severe pre-eclampsia.
  
For instance, one study found that women who engaged in any regular LTPA regardless of caloric expenditure were at decreased risk of preeclampsia, although another study did not find a significant association between the amount of time or intensity of physical activity in leisure time and the risk of preeclampsia or gestational hypertension (PUBMED:15466498; PUBMED:19916877). Another study suggested that women who performed regular LTPA had a reduced risk of pre-eclampsia and gestational hypertension, with the relative risks decreasing as the average time spent in LTPA increased (PUBMED:2592903).
However, another study documented an increased risk of severe pre-eclampsia with higher levels of LTPA (PUBMED:19055653). Similarly, a systematic review reported that high-intensity LTPA before and/or during pregnancy or more than 4 hours per week of LTPA may reduce the risk of pre-eclampsia, but also noted that one study found an elevated risk of severe pre-eclampsia with high amounts of LTPA (PUBMED:23836014).

Furthermore, a case-control study indicated that women who performed regular LTPA had a reduced risk of pre-eclampsia and gestational hypertension, with the relative risks decreasing as the average time spent in LTPA increased (PUBMED:26910608). A systematic review and meta-analysis of RCTs and cohort studies supported the promotion of LTPA in pregnancy as a strategy to improve maternal and child health, suggesting that participation in LTPA was associated with lower weight gain during pregnancy, lower likelihood of gestational diabetes mellitus (GDM), and lower risk of preterm delivery (PUBMED:27282925).

In conclusion, while some studies suggest that LTPA in early pregnancy may protect against pre-eclampsia, particularly at moderate intensities, the evidence is not consistent across all studies. Further research is needed to clarify the relationship between LTPA and pre-eclampsia risk, taking into account the intensity and duration of physical activity, as well as individual differences in susceptibility to the condition. Additionally, the potential influence of other factors such as prepregnancy BMI and socioeconomic status should be considered (PUBMED:32093248).} \\ 
  
\hline
\end{tabular}

\end{table*}

\end{document}